# ATM Cash demand forecasting in an Indian Bank with chaos and deep learning

Vangala Sarveswararao[1,2], Vadlamani Ravi[1*]

[1]Center of Excellence in Analytics,
Institute for Development and Research in Banking Technology
Castle Hills Road 1, Masab Tank, Hyderabad-500057 India.
[2]School of Computer and Information Sciences, University of Hyderabad-500046, India

sarveswararao.cs@gmail.com; rav_padma@yahoo.com

**Abstract.** This paper proposes to model chaos in the ATM cash withdrawal time series of a big Indian bank and forecast the withdrawals using deep learning methods. It also considers the importance of day-of-the-week and includes it as a dummy exogenous variable. We first modelled the chaos present in the withdrawal time series by reconstructing the state space of each series using the lag, and embedding dimension found using an auto-correlation function and Cao's method. This process converts the uni-variate time series into multi variate time series. The "day-of-the-week" is converted into seven features with the help of one-hot encoding. Then these seven features are augmented to the multivariate time series. For forecasting the future cash withdrawals, using algorithms namely ARIMA, random forest (RF), support vector regressor (SVR), multi-layer perceptron (MLP), group method of data handling (GMDH), general regression neural network (GRNN), long short term memory neural network and 1-dimensional convolutional neural network. We considered a daily cash withdrawals data set from an Indian commercial bank. After modelling chaos and adding exogenous features to the data set, we observed improvements in the forecasting for all models. Even though the random forest (RF) yielded better Symmetric Mean Absolute Percentage Error (SMAPE) value, deep learning algorithms, namely LSTM and 1D CNN, showed similar performance compared to RF, based on t-test.

## 1    Introduction

Financial forecasting usually involves prediction of time series such as stock market price, gold price, crude oil price, FOREX rate, interest rate and macroeconomic variables (eg. consumer price index). Phenomenal work has been reported in all these areas, where increasingly accurate models, both of stand-alone and hybrid types, were proposed. This area spans the statistical family of techniques such as simple linear regression, ridge regression, LASSO, multi-variate adaptive regression splines (MARS), Autoregressive Regressive Integrated Moving Average (ARIMA) and logistic regression on one hand and machine learning family of techniques such as different neural networks architectures, support vector regression, decision trees on the other [1]. Of late, deep learning neural network architectures have also been employed to solve these problems quite successfully. Many other types of time series also occur naturally while a bank performs its operations. Although not so well researched, one such problem pertains to offering services through Automated Teller Machines (ATM).

---

[*]Corresponding Author;

Accurate forecasting of cash withdrawals in ATMs is beneficial to both customers and banks. If banks load excess cash in ATMs than what is required, then they will lose out on the interest earned on the idle cash that would have otherwise been realized had that cash been invested elsewhere. On the other hand, if less than required cash is loaded in ATMs, then it results in ATMs going out of cash entailing severe customer dissatisfaction. Therefore, loading ATMs with optimal amount of cash is an important problem critical to the success of banking operations. Further, the frequency of replenishment of ATMs with cash is also an associated difficult problem. However, it turns out that ATM cash withdrawals forms a time series. If only accurate predictions are made to this time series, it will have far reaching ramifications to the cash dispensing, which among others, is traditionally treated as a yardstick to measure the overall success of the banking operations. Consequently, both issues (i.e. the amount of cash to be replenished and frequency of replenishment) mentioned above will be resolved in one go.

Identification and modeling of chaos present in a financial time series is of recent phenomenon (Pradeep and Ravi papers). As regards ATM cash withdrawals forecasting, there exists only one work (Venkatesh et al, 2014)[2] that exploited the power of identification and modeling of chaos before embarking on forecasting with a host of neural networks. Further, deep learning paradigm has not yet been fully exploited to solve this problem.

This paper precisely fills that gap. The main contributions of this paper are (i) first modeling the chaos present in the ATM withdrawals time series (ii) employing two popular deep learning architectures along with traditional machine learning techniques for forecasting purpose (iii) studying the effect of the *day-of-the-week* dummy variable as an exogenous variable on the accuracy of the forecasts (iv) for the first time, the study of forecasting the ATM cash withdrawals for the Indian Banks is conducted. This last aspect was studied earlier in Venkatesh et al (2014). In this study, we worked on a dataset taken from a big, well-known Indian Commercial bank, hereafter referred to as XYZ bank. The dataset contains daily cash withdrawals of 40 ATMs for two consecutive years.

Because it is a time series, we can use forecasting methods ranging from Autoregressive Regressive Integrated Moving Average (ARIMA) to complex models from the Machine learning & Deep Learning family. The popularity of this problem can be assessed by the fact that in 2005, Lancaster University conducted NN5 competition [3], which focused on forecasting daily cash withdrawals for 111 ATMs across the UK, and the participants proposed various methods to forecast cash withdrawals accurately.

The rest of the paper organized as follows: Section 2 presents the literature review; Section 3 presents in detail our proposed model; Section 4 presents the dataset description and evaluation metrics; Section 5 presents a discussion of the results and finally section 6 concludes the paper and presents future directions.

## 2 Literature survey

We first review the research works conducted on the dataset of NN5 Competition [3] as this competition is very much related to our work. The dataset in this competition contains two years of daily cash withdrawals for 111 ATMs in the UK. The task is to forecast ATM cash withdrawals for each ATM for the next 56 days. The best performing model from Andrawis et al. [4] is an ensemble of General Gaussian Regression, Neural network, and linear models and achieved a SMAPE of 18.95%. Venkatesh et al. [5] improved the forecast accuracy reported by Andrawis et al. [4] by clustering similar ATMs and applying a host of popular neural networks. In another work, Venkatesh et al [2] further improved the results by modelling chaos in the cash withdrawal time series and invoking the popular neural networks architectures. None of the studies included the effect of exogenous features such as 'day_of_week' dummy variable and whether it is weekday or weekend as these features have a significant effect on the forecast accuracy.

Venkatesh et al. [5] clustered the ATMs with the similar day of the week cash demand patterns and applied four neural networks namely General regression neural network (GRNN) [6], multi-layer feed-forward neural network (MLFF) [7], group method of data handling (GMDH), wavelet neural network (WNN) [8] and Auto-Regressive Integrated Moving Average (ARIMA). Out of their four networks, GRRN yielded an SMAPE of 18.44%, which is better than the result of Andrawis et al. [4]. Venkatesh et al. [2] found that chaos was present in the cash withdrawals data of NN5 Competition. In order to find and model chaos, they employed the TISEAN tool for calculating the optimal lag and embedding dimension of each series. They employed a chaotic approach to the forecasting and achieved a SMAPE of 14.71%, which was better than the results of Venkatesh et al. [5]. Later, Javanmard [9] employed fuzzy logic to forecast cash demand, and did not compare the proposed method with ML methods. Bhandari & Gill [10] proposed a hybrid of NN using Genetic Algorithm to predict cash demand and it outperformed NN with gradient descent. They did not mention the data used or comparison with other methods. Jadwal et al. [11] proposed a clustering method for ATMs but it is more or less, same as that of Venkatesh et al. [5]. Arabani & Komleh [12] proposed convolutional neural network (CNN) while Rafi et al. [13] employed Vector Auto Regression with Exogenous Variable model (VAR-MAX).

The use of machine learning and deep learning algorithms for forecasting time series became a common practice nowadays as they tend to outperform the traditional statistical models and can learn intricate patterns in the series. Pradeepkumar and Ravi [14][15]improved the FOREX rate predictions by modelling chaos before applying the hybrid of MLP + Evolutionary algorithms, Quantile regression random forest [16], and Multivariate Regression Splines [17]. Recurrent neural networks (RNNs) are commonly used deep learning algorithms for sequence prediction and they have gained popularity in solving complex sequential problems such as natural language understanding, speech recognition, and language translation. However, unlike MLP, the feedback loop of the recurrent cells addresses the temporal order of the sequences [18]. The commonly used RNN cells used for sequence prediction are Elman RNN cell (Elman, 1990), Gated Recurrent Unit (GRN) [19], and Long Short-term Memory cell [20]. Because of the vanishing gradient problem in the RNN cell, LSTM is preferred for modelling sequence problems. References [21], [22] and [23] successfully applied LSTMs to time-series forecasting applications. Bengio [24] proposed the 1- dimensional convolutional neural networks (ID-CNN)for time series forecasting, but they are inefficient in capturing long term dependency in the time series. Papadopoulos [25] proposed three layers dilated causal convolutions to time series forecasting based the WaveNet architecture for audio waveforms [26], which resulted in better performance compared to LSTM implementation. All these works on ATM cash withdrawals forecasting did not consider using exogenous features.

## 3 Overview of the techniques used

### 3.1 Chaos theory

Poincare proposed the theory of chaos during the 1800s, and later it was extended by Lorenz [27] in order to deal with complex non-linear systems [28]. A chaotic system is dynamic, deterministic and evolves from initial conditions, and trajectories can describe the system in the state space. The governing equations for a chaotic system are known ahead, so the state space is represented by phase space, which we can reconstruct from original series using delay time and embedding dimension [29] as shown in Eq. (1):

$$Y_i = (x_i, x_{i+\tau}, x_{i+2\tau}, \ldots, x_{i+(m-1)\tau}) \qquad (1)$$

Where τ is the time delay or the lag of the system and m is the embedding dimension to reconstruct the phase space. After reconstructing the phase space, the time series problem turns out to be a multi-input single-output (MISO) prediction problem, which can be modelled by methods ranging from linear models to deep neural networks.

### 3.2. Rosenstein's method

Rosenstein's algorithm [30] estimates the largest Lyapunov Exponent (λ) [31] from the given time series as shown as in the figure 1. If $\lambda \geq 0$ then chaos is present; otherwise chaos is absent in given time series. Fig. 1 depicts the method.

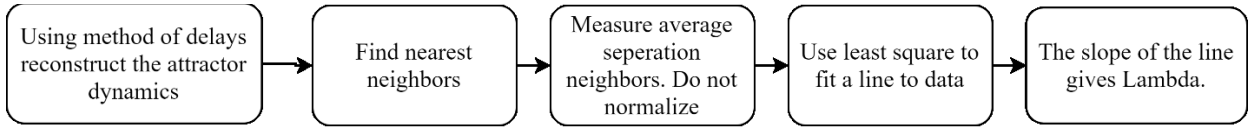

Fig. 1. Rosenstein's method to calculate λ

### 3.3. Cao's method

Cao proposed a method [32] to find out the minimum embedding dimension for a given times series. Let $X = (x_1, x_2, x_3, x_4, \ldots, x_N)$ be a time series. In the phase space, the time series can be reconstructed as time delay vectors as in Eq. (2):

$$Y_i = (x_i, x_{i+\tau}, x_{i+2\tau}, \ldots, x_{i+(m-1)\tau}) \qquad (2)$$

Where, $Y_i$ is the $i^{th}$ reconstructed vector, and τ is the time delay.

### 3.4 Methods employed in this study

#### 3.4.1. Autoregressive integrated moving average (ARIMA)

In time series analysis, non-seasonal ARIMA models are usually denoted as ARIMA(p, d, q) where p is autoregressive model order, d is the degree of differencing and q is the moving-average model order. These models are fitted to the given time series data either to predict future points in series or to understand the series better. ARIMA models are applied when the series shows non-stationarity, where differencing step (d) is applied to make the series stationary. The forecasting equation constructed as follows.

Let y denote $d^{th}$ difference of given time series $Y$.

If d=0, then $y_t = Y_t$. If d = 1, then $y_t = Y_t - Y_{t-1}$. If d = 2, then $y_t = (Y_t - Y_{t-1}) - (Y_{t-1} - Y_{t-2})$ etc.

Then the ARMA (p,q) model is written as follows:

$$\hat{y}_t = c + \varphi_1 y_{t-1} + \cdots + \varphi_1 y_{t-p} + \theta_1 e_{t-1} + \cdots + \theta_1 e_{t-q}$$

Here $\varphi's$ are autogressive parameters, $\theta's$ are moving average parameters and c is the intercept.

### 3.4.2. Support Vector Regression Machines (SVR)

Drucker et al. [33] proposed a version of Support Vector Machines [34] to deal with regression problems and the training phase of SVR involves minimizing the following equation.

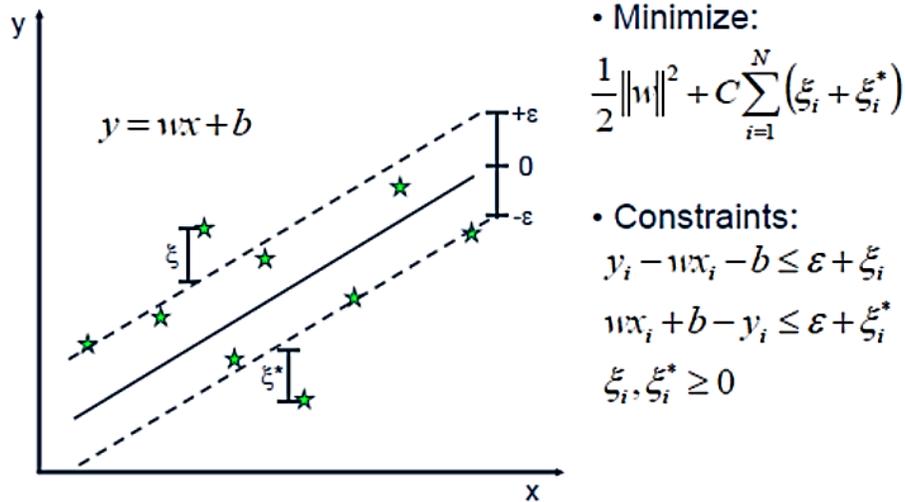

Fig. 2. Support Vector Regression

Where w = weight vector, N = the number of training examples, $y_i$ = target feature, b = bias, C = regularization constant and $\epsilon$ = margin of tolerance. Fig. 2 depicts the architecture of SVR.

### 3.4.3. Random forest (RF)

Random forests proposed by Ho [35] is an ensemble method for both classification and regression problems. RF works by constructing a multitude of decision trees during the training phase and outputs the target value by calculating the mean of the individual decision trees outputs.

*3.4.4. XG Boost*

Extreme Gradient Boosting (XGBoost) proposed by Chen [36] is a decision tree based ensemble algorithm that has been dominating Kaggle competitions and applied machine learning for tabular data. XGboost is robust implementation of gradient boosting trees [37], which are designed for performance and speed. Boosting is an ensemble method to construct a strong model using many weak models by adding models on top of each other, i.e. the mistakes or errors made by previous models are corrected by next predictor, until the training set is correctly predicted.

*3.4.5. Multi-layer perceptron (MLP)*

Multilayer perceptron is a feed forward artificial neural network that is trained by backpropagation algorithm. MLP consists of 3 layers namely an input layer, a hidden layer, and an output layer. Hidden and the output layers have sigmoid or logistic function as the non-linear activation function. It is too popular architecture to be discussed here in greater detail.

*3.4.6. Group method of data handling (GMDH)*

GMDH (Ivakhnenko, 1968) is a self-organised feed-forward network based on polynomial transfer functions, called Ivakhnenko polynomials. The coefficients of these functions are determined using least square regression. It is the best suitable ANN for dealing with inaccurate, noisy, or small datasets. It can be used to solve classification & regression problems and solve the problem of overfitting. It is the earliest proposed deep learning neural network architecture, without using the present day jargon, where nodes in the hidden layers are dropped if they are found to be not having sufficient predictive power. The GMDH algorithm builds regression equations of high order (e.g., 2 or 3) for each pair of input variables $x_i$ and $x_j$ (e.g., $y = a + bx_i + cx_j + dx_i^2 + ex_j^2 + fx_ix_j$ ). This is called an Ivakhnenko polynomial. The best input variables (polynomials) selected as per pre-specified selection criteria are retained and the next layer accepts these best variables to generate new input variables of higher-order (order of 4 if started with order 2). This process is continued as long as the prediction model does not satisfy the pre-specified error tolerance.

*3.4.7. General regression neural network (GRNN)*

Generalized regression neural network (GRNN) proposed by Specht [6] in 1991 is a feed forward neural network having roots in statistics. GRNN represents an advanced architecture in the neural networks that implements non-parametric regression with one-pass training. The topology of GRNN involves four layers of neurons, viz., input, pattern, summation, and the output in that order. The pattern layer comprises 'n' training neurons. The test sample is fed to the input layer, which has all the variables, while the pattern layer consists of all the training samples. The distance, di, between the training sample present as a neuron in the pattern layer and the data point from the test set used for regression, is used to figure out how well each training neuron in pattern layer can represent the feature space of the test sample, X. This probability density is calculated by the Gaussian activation function. Thus, the summation of the product of target value and the result of activation function for each neuron i.e., $\sum_{i=1}^{n} Y_i * e^{\left(\frac{-d_i^2}{2\sigma^2}\right)}$ forms the numerator term in the summation layer and the summation of the term $e^{\left(\frac{-d_i^2}{2\sigma^2}\right)}$ i.e., $\sum_{i=1}^{n} e^{\left(\frac{-d_i^2}{2\sigma^2}\right)}$ forms the denominator term in the summation layer. Thus, the summation layer has two neurons. Then, in the output layer containing one neuron, the prediction is calculated as $\hat{Y}(X) = \frac{\sum_{i=1}^{n} Y_i * e^{\left(\frac{-d_i^2}{2\sigma^2}\right)}}{\sum_{i=1}^{n} e^{\left(\frac{-d_i^2}{2\sigma^2}\right)}}$.

Here $d_i^2 = (X - X_i)^T * (X - X_i)$. X is a test sample and Xi is the neuron present in the pattern layer.

### 3.4.8 Long short term memory Neural Network

Long short term memory (LSTM) is a variation of recurrent neural network proposed by Hochreiter and Schmidhuber [20] in 1997 that is trained using back propagation through time and overcomes vanishing gradient problem. So, it can be used to construct a large network to deal with difficult sequence problems mostly in natural language processing. But, LSTM's can be used as a forecasting algorithm as time series is a sequence of data points recorded over time. Instead of a neuron, LSTM's contains memory block as nodes which are connected through layers. Each memory block has three types of gates in it. i) Forget Gate, which decides what information to throw away from the memory block, ii) Input Gate, that checks which values from the input to be used to update the memory block, iii) Output Gate, that decides what to output based on the input and memory of the block. Figure 3 depicts the architecture of an LSTM unit.

$$\tilde{c}^t = \tanh(W_c [a^{t-1}, x^t] + b_c)$$

$$Update\ Gate: \Gamma_u = \sigma(W_u [a^{t-1}, x^t] + b_u)$$

$$Forget\ Gate: \Gamma_f = \sigma(W_f [a^{t-1}, x^t] + b_f)$$

$$Output\ Gate: \Gamma_o = \sigma(W_o [a^{t-1}, x^t] + b_o)$$

$$c^t = \Gamma_u * \tilde{c}^t + \Gamma_f * c^{t-1}$$

$$a^t = \Gamma_o * \tanh c^t$$

Where $(b_c, b_u, b_f, b_o)$ are biases for cell, update gate, forget gate and output gate respectively. $(a^{t-1}, a^t)$ are previous cell and current activation values respectively. $(W_f, W_o, W_u)$ are weights for forget, output and update gates respectively. $(c^{t-1}, c^t)$ are previous and current cell memory values respectively and $\tilde{c}^t$ is temporary cell memory. Fig. 3 depicts the architecture of the LSTM unit.

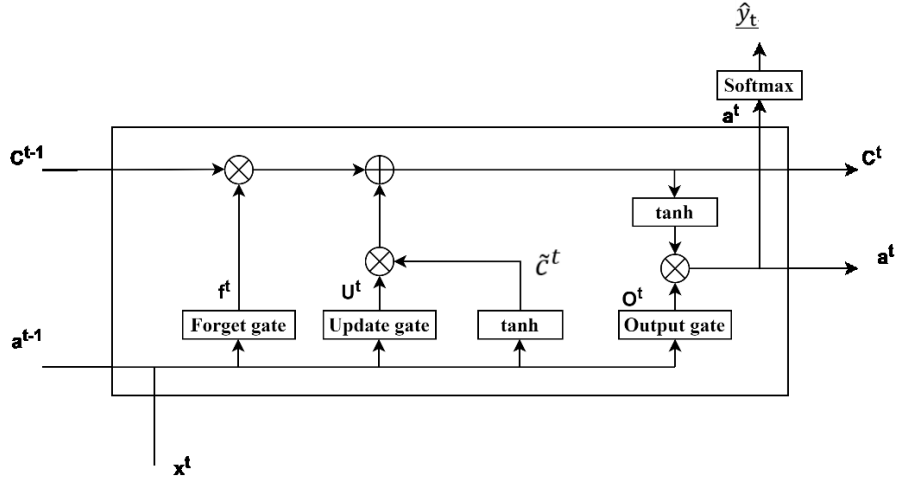

*Fig. 3. Architecture of LSTM Unit*

### 3.4.9 One dimensional Convolutional Neural Network (1D-CNN)

Convolutional Neural Network (CNN) wase proposed by LeCun [24] for robust image classification. CNN models take images as two-dimensional input by representing each image as its pixel values and colour channels and output its corresponding predicted class. Each CNN layer from left to right learns most basic to complex features of the image, and the final layer takes complex features and outputs a target class. The one-dimensional CNN (1D-CNN) also follows the same procedure as 2D CNN but learns features from sequences of data (eg. Time Series). 1D-CNN extracts features from the time-series data, and it uses them to forecast future values. CNN works the same way for any number of dimensions except the structure input data and filters vary. Fig. 4 depicts the architecture of the 1D-CNN.

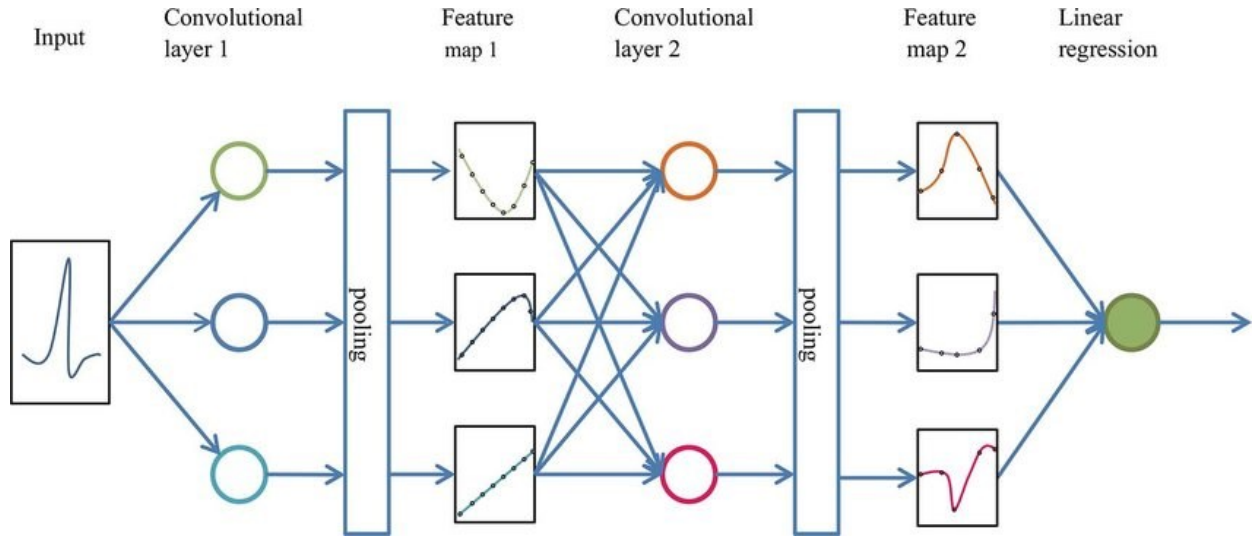

*Fig. 4. Architecture of 1D CNN*

# 4 Proposed Methodology

The proposed methodology (see Fig.1) first extracts the exogenous features such as a week of the day, weekday or weekend from the date index of each series and then it appends this additional information with the series data to forecast the predictions. We applied One-Hot-Encoding to the week of the day feature, which resulted in a total of 7 exogenous features, which when combined with two feature for weekend or weekday, resulted in a total of nine features. On Sunday and Saturday, cash withdrawals are, in general, high as it is a weekend. On the weekdays, withdrawals fluctuate according to the week of the day. We considered this extra information along with the series data while training machine learning and deep learning algorithms.

## 4.1 Description of the two-stage model

Let $Y = (y_1, y_2, y_3, \ldots, y_N)$ be a single ATM cash withdrawal series and the future cash demand is calculated using the following two-stage modelling.

*Stage 1: Chaos Modelling and Extracting features from the time index of the data set.*

- *Reconstructing the phase space*: Check series Y for the presence of chaos with the help of Lyapunov exponent value. If chaos is present, then reconstruct the phase space using lag/delay time ($\tau$) & embedding dimension (m) found using autocorrelation function and Cao's method respectively. Partition the Y into $Y_{Train} = \{y_t; t = \tau m + 1, \tau m + 2, \ldots, k\}$ and $Y_{Test} = \{y_t; t = k + 1, k + 2, \ldots, N\}$.

- *Extracting exogenous features from time index:* Extract "day_of_the_week" and "is_weekend" dummy features from the index and apply the one-hot-encoding to get a single feature for every week. Now we get seven extra features, namely, is_monday, is_thuesday, is_wednesday, is_thursday, is_friday, is_saturday and is_sunday. In addition, 'is_weekend' dummy variable results in 2 features after one-hot encoding. All these nine exogenous features will be appended to the chaos modelled data before invoking any ML / DL technique.

*Stage 2: Forecasting ATMs Cash Demand Using ML / DL Techniques*

In a year, first 11 months data of each ATM series is considered as training data, and the last 30 days are considered as the test set. The model consists of the reconstructed phase space and the exogenous variables as follows. Fig. 5 depicts the schematic of the proposed hybrid methodology.

$$\widehat{Y_t} = \alpha_0 + \alpha_1 y_{t-\tau} + \alpha_2 y_{t-2\tau} + \alpha_3 y_{t-3\tau} + \cdots + \alpha_m y_{t-\tau m} + \beta_1 y_{is\_monday} + \beta_2 y_{is\_tuesday} + \beta_3 y_{is\_wednesday} + \beta_4 y_{is\_thursday} + \beta_5 y_{is\_friday} + \beta_6 y_{is\_saturday} + \beta_7 y_{is\_sunday} + \beta_7 y_{is\_weekdend}$$

Where $\widehat{Y}$ = predicted value, $(\alpha_0, \alpha_1, \alpha_2, \ldots, \alpha_m)$ are coefficients for chaotic variables, $(\beta_0, \beta_1, \beta_2, \ldots, \beta_7)$ are coefficients for exogenous features

# 4 Dataset description and evaluation metrics

We collected the dataset from a well-known Indian commercial bank for this research study. The dataset contains daily cash withdrawals of the past two years, starting from 21-11-2017 to 20-11-2019 for 100 ATMs.

## 4.1 Dataset pre-processing

Because of the technical problems, a few ATMs were out of service in some days, which led to missing entries in the dataset. If the dataset contains too many missing entries, then the model may fit the imputed values. Therefore, we removed the data of 60 out of 100 ATMs as the number of missing entries are greater than 110 in those cases. We then imputed missing entries (withdrawals) for the rest 40 ATMs with a median withdrawal of the 40 ATMs.

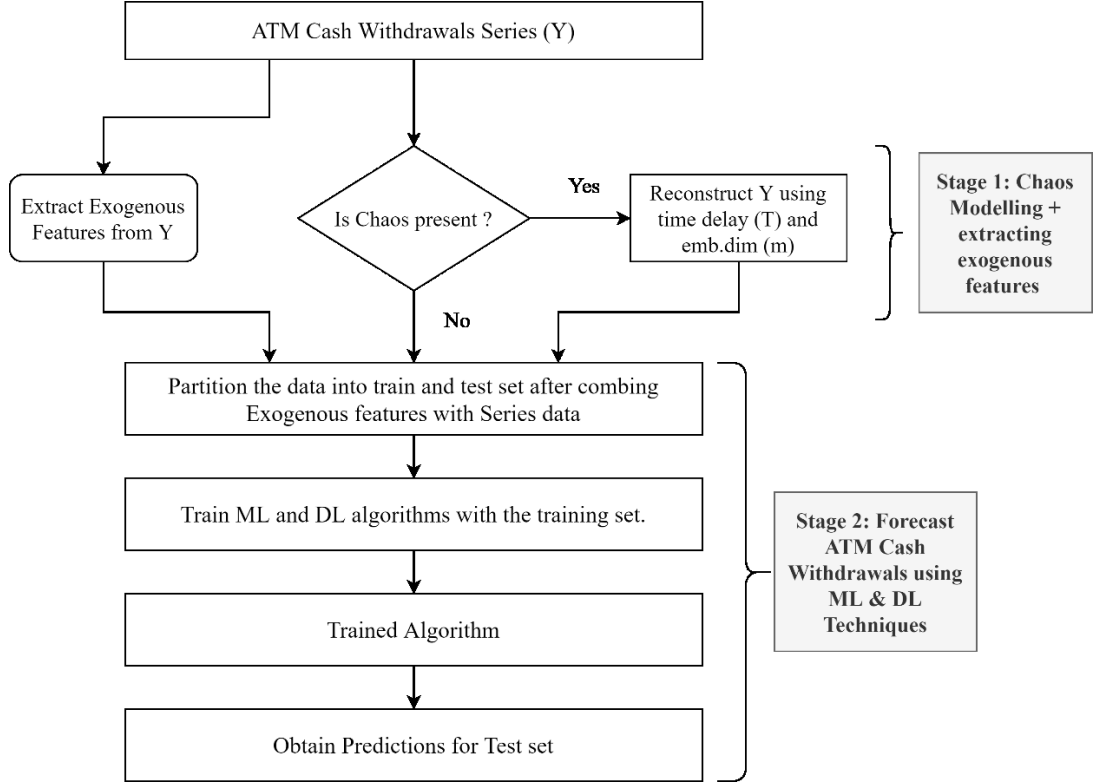

*Fig.5. Schematic of the Proposed methodology*

We used the Augmented Dickey-Fuller [38] test from the *statsmodel* library to check whether the series is stationary or not. The test indeed showed that the withdrawals in all the ATMs series are stationary. To deseasonalize the series, we employed the median-based deseasonalization method of [4]. We then computed the Lyapunov Exponent [31] using Rosenstein's method for all the ATMs to check for the presence of chaos in the cash withdrawals series and the exponent turns out to be higher than zero for all the ATMs, which confirms that chaos is present in the cash withdrawals series for all the ATMs. As the machine learning and deep learning algorithms require the data to be normalized, we performed min-max normalization on the dataset for all the ATMs.

Using Cao's method [32], minimum embedding dimension for all the ATMs turned out to be seven and using the partial auto-correlation function , the lag turned out to be one for each ATM series. We considered first 11 months data to be the training set and the last 30 days data to be the test set.

## 4.2 Evaluation measures

We considered SMPAE to measure the prediction performance of all the algorithms and across all the ATMs because each ATMs follow different ranges of cash withdrawals. In addition to SMAPE, we also considered Directional Symmetry [39] statistic and Theil's U Coefficient [40]. [41] and [42] for measuring forecasts in both directions and their closeness to original values.

### 4.2.1. Symmetric Mean Absolute Percentage Error (SMAPE)

SMAPE is used as a primary evaluation measure in time series as it is invariant to the scale of the series data and is defined as follows:

$$\text{SMAPE} = \frac{100\%}{n} \sum_{t=1}^{n} \frac{|\hat{Y}_t - Y_t|}{(|\hat{Y}_t| + |Y_t|)/2}$$

Where $\hat{Y}_t$ = forecasted value at time $t$, $Y_t$ is observed value at time $t$ and $n$ is the number of samples.

### 4.2.2. Theil's U Coefficient

Theil's U Coefficient is a relative accuracy measure that gives more weight to the larger errors, which can help eliminate models with large errors. If it is less than 1 then our forecast is better than random guessing whereas if it is greater than 1, then our forecast is worse than random guessing. It is defined as follows:

$$Theil's\ U\ statistic = \sqrt{\frac{\sum_{t=1}^{n-1}\left(\frac{\hat{y}_{t+1} - y_{t+1}}{y_t}\right)^2}{\sum_{t=1}^{n-1}\left(\frac{y_{t+1} - y_t}{y_t}\right)^2}}$$

### 4.2.3. Directional Symmetry (DS) Statistic

DS is a statistical measure of a model's performance in predicting the direction of change, positive/negative, of a time series from a one-time period to the next. It is defined as follows:

$$DS(t,\ \hat{t}) = \frac{100}{n-1} \sum_{i=2}^{n} d_i$$

$$d_i = \begin{cases} 1, & if\ (y_i - y_{i-1})(\hat{y}_i - \hat{y}_{i-1}) > 0 \\ 0, & otherwise \end{cases}$$

## 5 Results and Discussion

Table 1 presents the hyperparameters and their ranges used for all the algorithms across all 40ATMs. Table 2 presents the results of our study with and without exogenous features namely day-of-the-week and weekday/weekend. It clearly shows that chaos modelling and including exogenous features to the data set will results in improved forecasts. We considered ARIMA as a baseline model for our study as it is a robust statistical model and all ML & DL algorithms except GMDH & GRNN yielded superior performance than ARIMA model in terms of SMAPE. From Table 1, one can see that random forecast (RF) yielded the best performance in terms of SMAPE and Dstat with or without using exogenous features. But, according to Theils U statistic, LSTM turned out to be the best when exogenous features were considered while MLP is the best when exogenous features were not considered. Therefore, in order to have a categorical inference from the study, we performed paired t-test on the mean SMAPE of the top three models on the test data at 40+40-2=78 degrees of freedom at a 5% level of significance to check whether they are statistically performing the same or not. The results are presented in Table 3. It turned out that the LSTM and 1D-CNN produced similar performance when compared with random forest. However, 1D-CNN turned out to be significant than the LSTM with or without exogenous dummy variables.

**Table 1.** Hyperparameters for all the algorithms

| Model | Hyperparameters |
|---|---|
| ARIMA | (p, d, q) = (7,0,0) |
| MLP | 2 layers, nodes in layer 1 = range(1, 64, step=4), nodes in layer 2 = range(1, 8), lr=0.1, momentum=0.9 |
| XGBOOST | n_estimators = range(100, 2000, step=100), max_depth = range(2, 10) |
| RF | n_estimators = range(5, 50), depth = range(4, 15) |
| SVR | C = [0.01, 1, 10, 100, 200, 300, 400, 500, 1000], gamma = [0.0001, 0.0003, 0.0006, 0.001, 0.005, 0.01, 0.03, 0.06, 0.09], Kernel = ['linear', 'rbf', 'sigmoid', 'poly'] |
| GRNN | Standard deviation = [0.1, 0.2, 0.3, 0.4, 0.5, 0.6, 0.7], smoothing factor=(0.01, 0.09) |
| GMDH | Max_layer_count = 50, ref_function_types = (linear, linear_cov, quardratic, cubic), alpha=0.5, n_jobs=4, admix_features = True. |
| 1D CNN | N_filters = range(5, 70, step=5), epochs = range(300, 2000, step=200), kernel_size=2, activation= 'relu', dense layer = range(2, 8), optimizer= 'adam', loss= 'mse'. |
| LSTM | Nodes range = (2, 10), epochs = range(300, 1000, step=100), optimizer= "adam". Lr=0.01. |

Figures 6 depicts the Boxplots for each algorithm with and without exogenous features and it clearly shows adding exogenous features leads to an increase in the interquartile range (IQR) for each algorithm while lowering the median SMAPE. Here, we have to two choices, (i) Do not consider extra features if getting smaller IQR is the objective and be ready to scarifies the improvements in media SMAPE of all methods. (ii) Consider extra features to get improvement in the median SMAPE coupled with an increment in the IQR. Thus there is a trade-off. So, the decision to include these features is left to the domain expert. The ML models did not perform well as they are not able to effectively learn sequential information like LSTM or 1D CNN. However, random forest gained the top spot with help of group knowledge of the trees it has been trained on. The supplementary file provides 40 figures, ATM-wise, depicting the SMAPE values obtained by random forest, 1D-CNN and LSTM in the case of including the exogenous variables.

**Table 2.** Results of the algorithms

|  | Without Exogenous features | | | | | | With Exogenous features | | | | | |
|---|---|---|---|---|---|---|---|---|---|---|---|---|
|  | SMAPE | | Dstat | | Theils U | | SMAPE | | Dstat | | Theils U | |
| **Model** | **Mean** | **Median** | **Mean** | **Median** | **Mean** | **Median** | **Mean** | **Median** | **Mean** | **Median** | **Mean** | **Median** |
| ARIMA | 24.35(2.77) | 24.91 | 51.72 (7.55) | 53.44 | 0.38 (0.15) | 0.33 | 25.40 (2.94) | 25.04 | 56.61 (5.9) | 62.07 | 0.41 (0.17) | 0.45 |
| MLP | 23.69 (2.80) | 24.20 | 52.75 (7.51) | 51.72 | 0.35 (0.17) | **0.32** | 22.66 (2.90) | 23.08 | 62.87 (8.89) | 62.07 | 0.36 (0.14) | 0.37 |
| XGBOOST | 23.89 (3.36) | 24.06 | 59.65 (8.02) | 58.62 | 0.41 (0.18) | 0.38 | 23.63 (3.50) | 23.54 | 60.81 (4.83) | 58.62 | 0.39 (0.16) | 0.39 |
| RF | **22.39 (3.03)** | **22.14** | **61.61 (6.53)** | **62.07** | 0.39 (0.17) | 0.37 | **21.48 (3.34)** | **21.40** | **63.45 (6.51)** | 62.07 | 0.34 (0.13) | **0.33** |
| SVR | 23.63 (2.77) | 24.08 | 53.49 (5.95) | 51.72 | 0.36 (0.18) | 0.36 | 22.81 (2.95) | 22.57 | 62.76 (5.80) | 62.07 | 0.38 (0.16) | 0.38 |
| GRNN | 24.61 (2.73) | 25.20 | 52.64 (7.66) | 51.72 | 0.41 (0.19) | 0.40 | 24.06 (2.86) | 23.98 | 61.61 (5.61) | 62.07 | 0.39 (0.17) | 0.42 |
| GMDH | 24.64 (2.66) | 25.15 | 53.34 (7.00) | 51.72 | 0.40 (0.18) | 0.36 | 23.91 (2.71) | 24.04 | 54.94 (5.55) | 55.17 | 0.38 (0.16) | 0.37 |
| 1D CNN | 23.09 (2.53) | 23.60 | 61.23 (6.98) | 59.41 | 0.36 (0.20) | 0.33 | 22.36 (2.72) | 22.53 | 62.71 (6.82) | **62.96** | 0.34 (0.19) | 0.35 |
| LSTM | 23.14 (2.47) | 23.60 | 60.57 (7.12) | 58.92 | **0.35 (0.16)** | 0.32 | 22.56 (2.63) | 23.16 | 62.13 (6.58) | 60.51 | **0.32 (0.16)** | 0.34 |

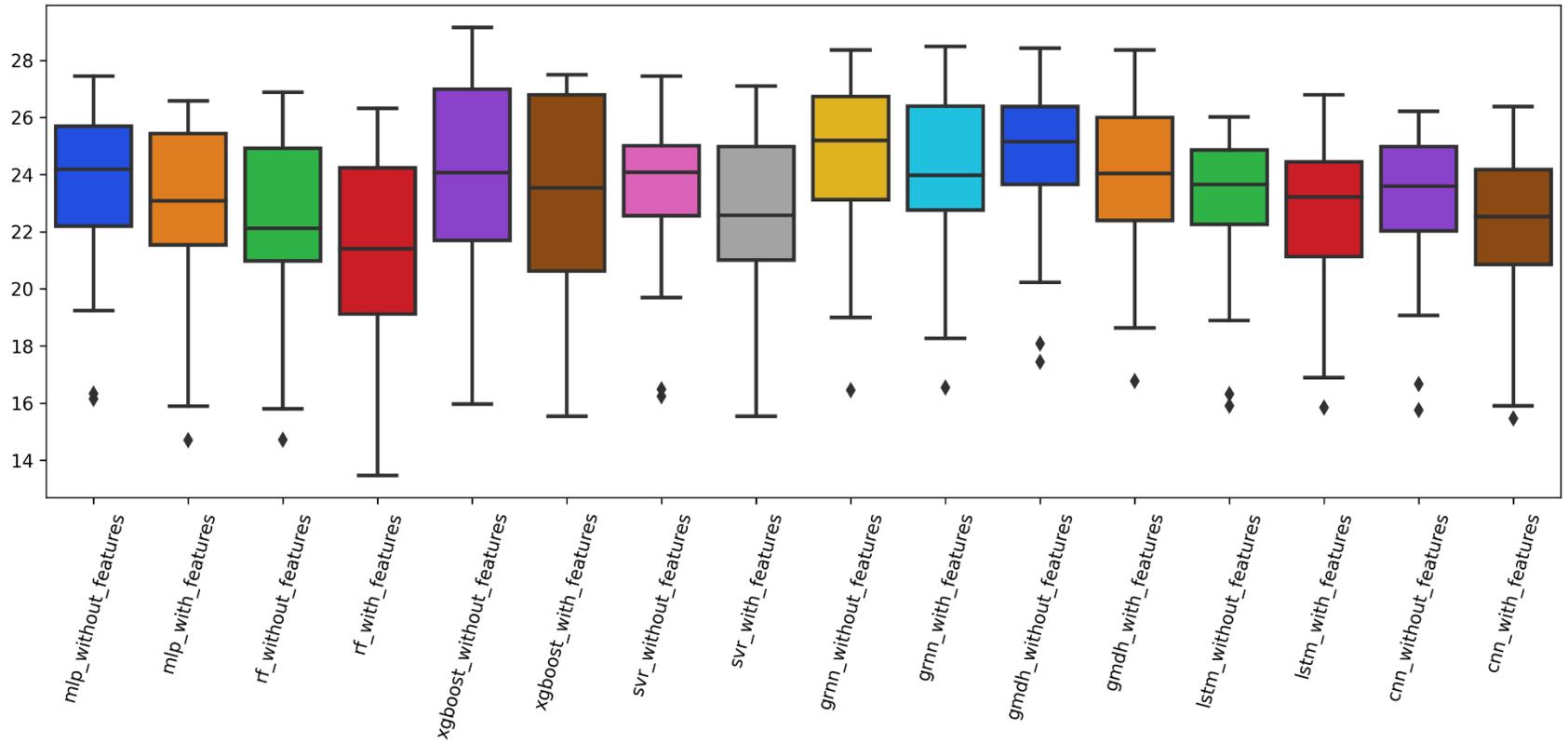

Fig. 6. Box plots for both ML and DL Algorithms.

**Table 3.** T-test for comparing top three performing models for ATM

| T Test at 0.05 Significance Level | P Value |
|---|---|
| Random Forest    Vs    LSTM | 0.009 |
| Random Forest    Vs    1D-CNN | 0.031 |
| LSTM Vs 1D-CNN | 0.289 |

# 6      Conclusion and Future Directions

In this study, we developed a hybrid model comprising chaos modeling and deep learning for forecasting cash withdrawals in ATMs of an Indian bank. This is an important operational problem for any bank, which would like to optimize it ATM replenishment activities. We also studied the impact of the exogenous features such as day-of-the-week and week-day/weekend dummy variables on the time series predictions instead of just using the raw time series data. All the forecasting algorithms showed an improvement in the SMAPE after adding these features, and also deep learning algorithms yielded statistically similar SMAPE when compared to the best performing machine learning algorithm in this case, namely random forest.

In the future, we would like to build a hybrid of Deep Learning and Machine Learning algorithms to leverage their advantages. We would like to see how transfer learning can be applied to time series, which might improve the performance on smaller time-series datasets.